\documentclass[letterpaper]{article} % DO NOT CHANGE THIS
\usepackage{aaai2026}  % DO NOT CHANGE THIS
\usepackage{times}  % DO NOT CHANGE THIS
\usepackage{helvet}  % DO NOT CHANGE THIS
\usepackage{courier}  % DO NOT CHANGE THIS
\usepackage[hyphens]{url}  % DO NOT CHANGE THIS
\usepackage{graphicx} % DO NOT CHANGE THIS
\usepackage{siunitx}
\usepackage{booktabs}
\usepackage{amsmath}
\usepackage{xcolor}
\usepackage{natbib}
\usepackage{natbib}  % DO NOT CHANGE THIS AND DO NOT ADD ANY OPTIONS TO IT
\usepackage{caption} % DO NOT CHANGE THIS AND DO NOT ADD ANY OPTIONS TO IT
\newcommand{\project}[0]{\pi}
\newcommand{\meanW}[0]{\boldsymbol{\mu}_{W}}

\newcommand{\meanI}[0]{\boldsymbol{\mu}_{I}}
\newcommand{\covW}[0]{\boldsymbol{\Sigma}_{W}}
\newcommand{\covI}[0]{\boldsymbol{\Sigma}_{I}}
\newcommand{\cam}[0]{\boldsymbol{T}}
\newcommand{\camCW}[0]{\cam_{CW}}

\newcommand{\gaussians}[0]{\mathcal{G}}

\newcommand{\SE}[1]{\boldsymbol{SE}(#1)}

\newcommand{\matJ}[0]{\mathbf{J}}
\newcommand{\matW}[0]{\mathbf{W}}

\usepackage{diagbox}
\usepackage{bbding}
\usepackage{algorithm}
\usepackage{algpseudocode}
\usepackage{subfigure}
\usepackage{multirow}
\usepackage{makecell}
\usepackage{graphicx}
\usepackage{amsmath}
\usepackage{amssymb}
\usepackage{booktabs}
\usepackage{xcolor}
\usepackage{colortbl}
\usepackage{upgreek}
\nocopyright
\usepackage[most]{tcolorbox} % 引入 tcolorbox 宏包

\newtcolorbox{promptbox}{
    colback=cyan!5!white,        % 更柔和的淡蓝背景（比 blue!10 更贴近图中效果）
    colframe=cyan!55!black,      % 浅蓝灰色边框
    fonttitle=\bfseries,     
    coltitle=white,          
    title=Prompt:,           
    arc=1mm,                 
    boxrule=1pt, 
    width=7.5cm, % 设置固定宽度为 10 厘米
}

\usepackage{newfloat}
\usepackage{listings}
\DeclareCaptionStyle{ruled}{labelfont=normalfont,labelsep=colon,strut=off} % DO NOT CHANGE THIS
\floatstyle{ruled}
\newfloat{listing}{tb}{lst}{}
\floatname{listing}{Listing}
\title{UW-3DGS: Underwater 3D Reconstruction with Physics-Aware Gaussian Splatting}
\author {
    Wenpeng Xing\textsuperscript{\rm 1},
    Jie Chen\textsuperscript{\rm 2},
    Zaifeng Yang\textsuperscript{\rm 3},
    Changting Lin\textsuperscript{\rm 4},
    Jianfeng Dong\textsuperscript{\rm 5},
    Chaochao Chen\textsuperscript{\rm 1},
    Xun Zhou\textsuperscript{\rm 6},
    Meng Han\textsuperscript{\rm 1}
}

\affiliations {
    \textsuperscript{\rm 1}Zhejiang University \quad
    \textsuperscript{\rm 2}Hong Kong Baptist University, Hong Kong SAR\quad
    \textsuperscript{\rm 3}A*STAR, Singapore\quad
    \textsuperscript{\rm 4}Binjiang Institute of Zhejiang University\quad
    \textsuperscript{\rm 5}Zhejiang Gongshang University\quad
    \textsuperscript{\rm 6}Harbin Institute of Technology\\
}

\usepackage{bibentry}

\begin{document}

\maketitle

\begin{abstract}
\begin{quote}
Underwater 3D scene reconstruction faces severe challenges from light absorption, scattering, and turbidity, which degrade geometry and color fidelity in traditional methods like Neural Radiance Fields (NeRF). While NeRF extensions such as SeaThru-NeRF incorporate physics-based models, their MLP reliance limits efficiency and spatial resolution in hazy environments.
We introduce UW-3DGS, a novel framework adapting 3D Gaussian Splatting (3DGS) for robust underwater reconstruction. Key innovations include: (1) a plug-and-play learnable underwater image formation module using voxel-based regression for spatially varying attenuation and backscatter; and (2) a Physics-Aware Uncertainty Pruning (PAUP) branch that adaptively removes noisy floating Gaussians via uncertainty scoring, ensuring artifact-free geometry.
The pipeline operates in training and rendering stages. During training, noisy Gaussians are optimized end-to-end with underwater parameters, guided by PAUP pruning and scattering modeling. In rendering, refined Gaussians produce clean Unattenuated Radiance Images (URIs) free from media effects, while learned physics enable realistic Underwater Images (UWIs) with accurate light transport.
Experiments on SeaThru-NeRF and UWBundle datasets show superior performance, achieving PSNR of 27.604, SSIM of 0.868, and LPIPS of 0.104 on SeaThru-NeRF, with ~65\% reduction in floating artifacts.

\end{quote}

\end{abstract}

\section{Introduction}

Accurate 3D scene reconstruction is fundamental to applications ranging from immersive virtual environments to marine exploration and underwater archaeology. However, underwater imaging remains challenging due to depth-dependent light absorption, scattering, and turbidity, which degrade color fidelity and geometry. Traditional methods and neural volumetric models like NeRF~\cite{mildenhall2020nerf} struggle in such conditions, as they assume clear media and cannot disentangle complex underwater light transport.

Recent extensions such as SeaThru-NeRF~\cite{levy2023seathru} incorporate underwater image formation models, but their reliance on MLPs limits spatial resolution and hampers accurate geometry recovery in scattering-dominated scenes.

To address these limitations, we propose \textbf{UW-3DGS}, a novel framework that adapts 3D Gaussian Splatting (3DGS)~\cite{kerbl3Dgaussians} for underwater 3D reconstruction. Our method integrates two key innovations: (1) a \textit{learnable underwater image formation module} that simulates wavelength-dependent attenuation and backscatter via voxel-based parameter regression, and (2) a \textit{Physics-Aware Uncertainty Pruning (PAUP) Branch} that removes floating Gaussians based on uncertainty scores, enhancing geometric fidelity.

UW-3DGS operates in two stages: during the \textbf{Training Stage}, noisy 3D Gaussians are jointly optimized with underwater parameters using end-to-end supervision from real underwater images. The PAUP branch prunes unreliable Gaussians, while the image formation module learns spatially varying scattering effects. In the \textbf{Rendering Stage}, the refined 3D Gaussians, optimized through the training process, are directly rasterized to produce clean, water-independent Unattenuated Radiance Images (URIs), capturing the intrinsic scene radiance free from scattering effects. Meanwhile, the learned physics parameters from the learnable underwater image formation module are applied to these Gaussians to generate realistic Underwater Images (UWIs), incorporating accurate light attenuation and backscatter as simulated by the module. This dual-output capability, driven by the module's spatially adaptive modeling, supports high-fidelity novel view synthesis and facilitates downstream visual tasks such as marine mapping, ecological analysis, and autonomous underwater navigation.

Extensive experiments on real-world datasets demonstrate UW-3DGS’s superior performance. On the SeaThru-NeRF dataset, it achieves a PSNR of 27.604, SSIM of 0.868, and LPIPS of 0.104. Our method reduces floating artifacts by ~65\% and preserves fine-grained structures such as coral textures and seabed contours, outperforming prior approaches in both geometric accuracy and visual realism.

Contributions include:
\begin{itemize}
\item Pioneering integration of learnable underwater physics into 3DGS, enabling exceptional URI and UWI quality.
\item Novel PAUP Branch for uncertainty-driven pruning, yielding artifact-free underwater geometry.
\item Demonstrated advancements in reconstruction accuracy on challenging underwater datasets, advancing practical utility.
\end{itemize}

\section{Related Work}

\begin{figure*}[h]
    \centering
    \includegraphics[width=\linewidth]{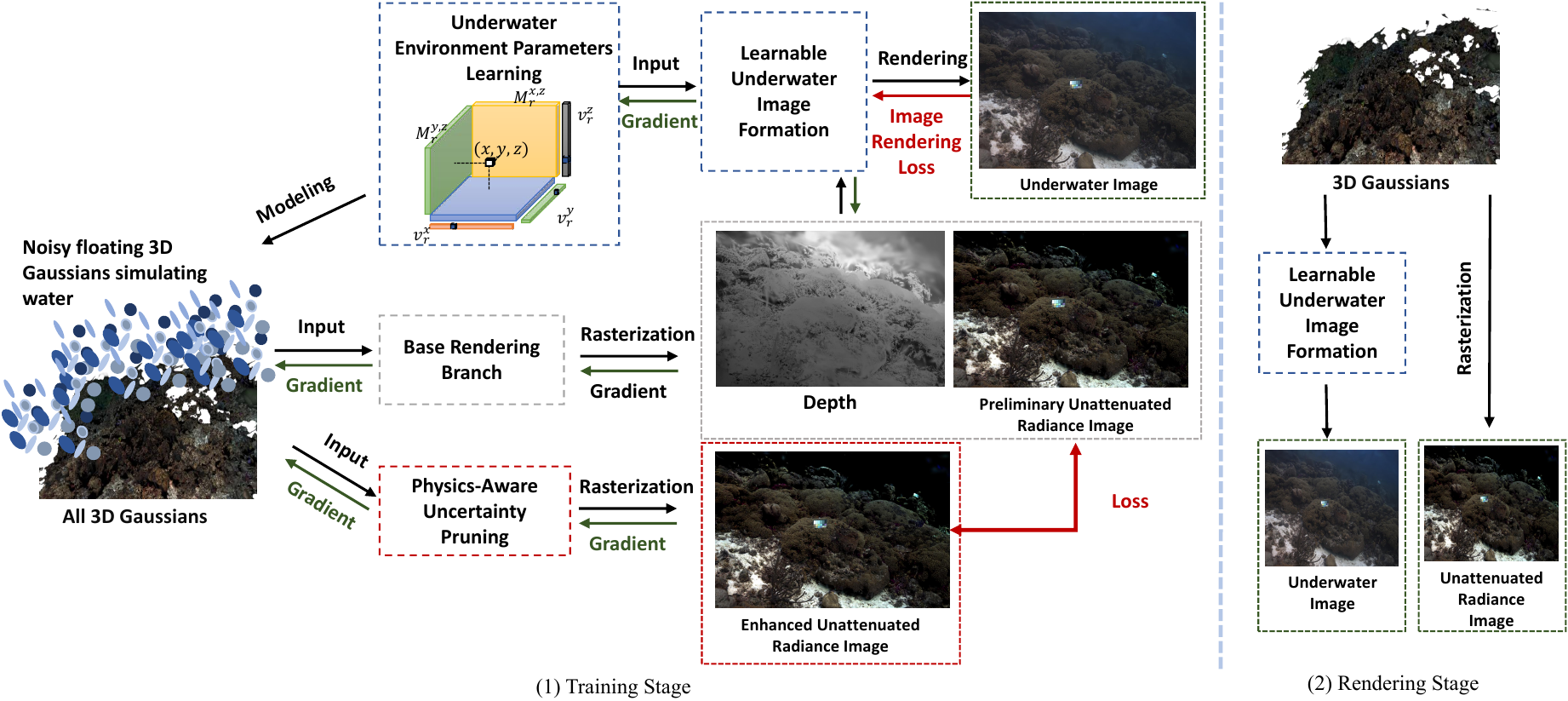}
    \caption{Architecture of UW-3DGS. In the training Stage, the Base Rendering Branch generates the preliminary Unattenuated Radiance Image (URI) and depth, the PAUP Branch prunes floating Gaussians using PAPSL, and the Learnable Underwater Image Formation Module applies scattering effects to produce the underwater image, all optimized end-to-end. In the Rendering Stage, refined Gaussians yield clean URIs, while learnable underwater image formation module enable realistic UWIs with accurate attenuation and backscatter.
    }
    
    \label{fig:enter-label}
\end{figure*}

Neural Radiance Fields (NeRF)~\cite{mildenhall2020nerf} have revolutionized 3D scene reconstruction and novel view synthesis, demonstrating exceptional fidelity. Their versatility extends to 2D image enhancement tasks, including denoising~\cite{pearl2022nan}, deblurring~\cite{ma2022deblur}, super-resolution~\cite{wang2022nerf}, and low-light enhancement~\cite{mildenhall2022nerf}, as well as robotics applications such as Simultaneous Localization and Mapping (SLAM)~\cite{rosinol2023nerf, yan2023gs} and robotic grasping~\cite{kerr2023evo}.

\subsection{Participating Media and Underwater NeRF}
Adaptations of NeRF to participating media, especially underwater environments, have gained traction to address light scattering and absorption challenges. Early works like SeaThru-NeRF~\cite{levy2023seathru} and WaterNeRF~\cite{sethuraman2023waternerf} incorporate physics-based image formation models into NeRF's volumetric framework, using MLPs to simulate attenuation and backscattering. WaterHE-NeRF~\cite{zhou2023waterhe} leverages histogram equalization for pseudo-ground truth supervision, while Dehaze-NeRF~\cite{chen2023dehazenerf} applies atmospheric scattering models to hazy scenes.
Recent advancements extend these foundations. NeuroPump~\cite{guo2024neuropump} introduces self-supervised geometric and color rectification to "pump out" water effects in NeRF reconstructions. AquaNeRF~\cite{gough2025aquanerf} proposes an MLP-based scheme for distractor-aware rendering. 
Despite these innovations, MLP reliance often results in prolonged training times, and validations are frequently limited to controlled settings. Emerging methods increasingly incorporate 3DGS for efficiency, as discussed below.

\subsection{Underwater 3D Gaussian Splatting}
Building on 3DGS's efficiency, recent works adapt it for underwater reconstruction to mitigate scattering-induced artifacts. SeaSplat~\cite{yang2024seasplat} enables real-time rendering by combining 3DGS with a physically grounded image formation model, disentangling medium effects from scene radiance. WaterSplatting~\cite{li2024watersplatting} fuses volumetric rendering with 3DGS, incorporating distractor-aware mechanisms for enhanced clarity in turbid waters.
Further developments include UW-GS~\cite{wang2025uw}, a distractor-aware variant with physics-based density control, and RUSplatting~\cite{jiang2025rusplatting}, which bolsters robustness for sparse-view scenarios through improved Gaussian optimization. Water-Adapted 3DGS~\cite{fan2025water} introduces complexity-adaptive point distribution and depth-based multi-scale rendering for precise scene recovery. For dynamic environments, UDR-GS~\cite{du2024udr} extends to 4D Gaussians, addressing temporal variations in underwater light propagation. 
These methods demonstrate improved scalability for open-ocean applications, though challenges in handling extreme turbidity and real-time deployment persist. In contrast, our UW-3DGS distinguishes itself by introducing a physics-aware uncertainty pruning branch to adaptively suppress floating Gaussians and a plug-and-play learnable underwater image formation module with voxel-based regression, enabling superior media-free reconstruction and end-to-end optimization.

\subsection{Light Propagation in Scattering Media}
Fundamental research on light propagation in scattering media underpins these advancements. SeaThru models~\cite{akkaynak2018revised, akkaynak2019sea, akkaynak2017space} emphasize wavelength-dependent parameters in underwater optics. Recent reviews by Yang et al.~\cite{yang2019depth} cover monocular restoration techniques, while Sharma et al.~\cite{sharma2021single} survey deep learning-based defogging.  For a thorough overview, consult the referenced surveys.

\section{Preliminaries}

\subsection{3D Gaussian Splatting (3DGS)}

UW-3DGS builds upon 3DGS~\cite{kerbl3Dgaussians}, which represents scenes as sets of anisotropic Gaussians $\{\gaussians^i \mid i \in [1, N]\}$ for efficient tile-based rasterization and real-time rendering. This representation is particularly promising for underwater scenes, where traditional volumetric methods like NeRF struggle with scattering-induced artifacts.

Gaussians are initialized from sparse point clouds generated by Structure-from-Motion (SfM) tools such as COLMAP~\cite{schoenberger2016sfm}. Each Gaussian $\gaussians^i$ includes view-dependent color $c^i$ (modeled via spherical harmonics) and opacity $\alpha^i$. The position and shape are defined by mean $\meanW^i$ and covariance $\covW^i$ in world space, decomposed as:
\begin{equation}
    \covW = RSS^\mathsf{T}R^\mathsf{T},
\end{equation}
where $S$ is the scaling matrix and $R$ is the rotation matrix.

During rasterization, 3D Gaussians are projected to 2D via:
\begin{equation}
    \meanI = \project ( \camCW  \meanW )~, \covI = \matJ\matW\covW \matW^T \matJ^T,
    \label{eqn:mean_cov_w2i}
\end{equation}
where $\project(\cdot)$ is the projection operation, $\matJ$ is the Jacobian of the affine approximation, $\camCW \in \SE{3}$ is the camera pose, and $\matW$ is the viewing transformation.

Pixel colors $\mathbf{C}$ are computed through alpha blending:
\begin{equation}
\mathbf{C} = \sum_{i\in N} \mathbf{c}_i \alpha_i \prod_{j=1}^{i-1} (1-\alpha_j),
\end{equation}
accumulating color contributions modulated by opacity and transmittance $T = \prod_{j=1}^{i-1} (1-\alpha_j)$. This fully differentiable formulation enables gradient-based optimization of Gaussian parameters, enhancing scene representation and image fidelity. 

\subsection{Underwater Image Formation Model}

To simulate light propagation in scattering media, UW-3DGS adopts a physically grounded underwater image formation model, enabling realistic rendering and geometry-aware restoration.

Early models~\cite{jaffe1990computer, schechner2005recovery} express observed intensity $I(x)$ at pixel $x$ as:
\begin{equation}\label{eq:img_formation_model_1}
    I(x) = D(x) + B(x),
\end{equation}
where $D(x)$ is the attenuated direct signal from the scene point, and $B(x)$ is the backscattered light by water particles. Color degradation primarily stems from wavelength-dependent attenuation in $D(x)$, while $B(x)$ reduces contrast via a veiling effect.

We adopt a revised formulation~\cite{akkaynak2018revised} for precise physical modeling:
\begin{equation}
I = 
\underbrace{J \cdot e^{-\beta^D \cdot z}}_{\text{direct transmission}} 
+ 
\underbrace{B^\infty \cdot (1 - e^{-\beta^B \cdot z})}_{\text{backscattering}},
\label{eq.scattering_1}
\end{equation}
where $J$ is the intrinsic scene radiance, $\beta^D$ and $\beta^B$ are attenuation coefficients for direct and backscatter signals, $z$ is the scene depth, and $B^\infty$ is the far-field veiling light. The first term models exponential decay due to absorption and scattering, while the second captures accumulated backscatter increasing with depth.

\section{Approach}

Directly applying 3D Gaussian Splatting (3DGS)~\cite{kerbl3Dgaussians} to underwater imagery yields noisy, floating Gaussians due to unmodeled light absorption and scattering, causing distorted geometry, inaccurate colors, and loss of details like seabed contours—limiting marine exploration, robotics, and ecological applications.

We propose \textbf{UW-3DGS}, an end-to-end framework for underwater 3D reconstruction that disentangles scattering effects from intrinsic scene properties, enhancing geometry accuracy and novel view synthesis. It comprises three integrated components:

\begin{enumerate}
\item \textbf{Base Rendering Branch:} An adapted 3DGS pipeline generating initial depth maps and unattenuated radiance images.
\item \textbf{Physics-Aware Uncertainty Pruning (PAUP):} An auxiliary branch pruning noisy Gaussians using voxel-wise uncertainty for improved consistency.
\item \textbf{Learnable Underwater Image Formation Module:} A physics-based model simulating light propagation with spatially varying parameters via voxel regression guided by PAUP.
\end{enumerate}

The framework operates in two stages (Figure.~\ref{fig:enter-label}):

\begin{enumerate}
\item {Training Stage.} Starting with noisy 3D Gaussians, the Base Rendering Branch produces initial radiance images and depth maps, while PAUP prunes unreliable Gaussians based on uncertainty. These feed into the Learnable Underwater Image Formation Module to model attenuation and scattering with learned parameters. End-to-end training minimizes rendering errors against ground truth using gradient-based losses.

\item  {Rendering Stage.} Post-training, refined Gaussians and learned parameters render high-fidelity unattenuated radiance images and underwater views, preserving details efficiently for novel view synthesis, visualization, and navigation.
\end{enumerate}

\subsection{Base Rendering Branch}

The Base Rendering Branch adapts the 3DGS pipeline~\cite{kerbl3Dgaussians} to generate initial representations, including the preliminary Unattenuated Radiance Image (URI) \(\hat{I}_{\text{UR}}\) and depth map \(z\), which serve as inputs for the PAUP and underwater image formation modules. Unlike standard 3DGS, we incorporate underwater-specific optimizations to mitigate scattering-induced artifacts.

Rendering follows 3DGS rasterization (Eq.~\eqref{eqn:mean_cov_w2i}), producing:

\begin{equation}
    \hat{I}_{\text{UR}} = \sum_{i\in N} \mathbf{c}_i \alpha_i \prod_{j=1}^{i-1} (1-\alpha_j).
\end{equation}

Depth is computed with uncertainty weighting from the PAUP branch:

\begin{equation}
    z = \sum_{i \in N} z_i \alpha_i (1 - U^i) \prod_{j=1}^{i-1} (1 - \alpha_j),
    \label{eq:base_depth}
\end{equation}

where \(z_i\) is the Gaussian’s depth and \(U^i\) is the uncertainty component from PAUP, prioritizing low-uncertainty contributions to mitigate scattering-induced depth errors.

\subsection{Physics-Aware Uncertainty Pruning Branch}

\label{sec:paup}

To suppress noisy floating Gaussians and enhance reconstruction quality, we introduce the Physics-Aware Uncertainty Pruning (PAUP) Branch, operating in parallel with the Base Rendering Branch. This branch uses a physics-aware uncertainty score (PUS) to guide adaptive pruning and provide uncertainty feedback for parameter regression in the underwater module.

For the \(i\)-th Gaussian \(\gaussians^i\), PUS is computed as:

\begin{equation}
    \text{PUS}^i = w_u \cdot U^i + w_p \cdot P^i,
\end{equation}

where \(w_u, w_p\) are learnable weights initialized to 0.5. The uncertainty component \(U^i\) captures rendering instability:

\begin{equation}
    U^i = w_\alpha \cdot \text{Var}_{\text{views}}(\alpha^i_{\text{eff}, k}) + w_c \cdot \text{Var}_{\text{views}}(\mathbf{c}^i(\mathbf{v}_k)),
\end{equation}

with \(w_\alpha = 0.4\), \(w_c = 0.6\). The effective opacity is:

\begin{equation}
    \alpha^i_{\text{eff}, k} = \alpha^i \cdot \prod_{j=1}^{i-1} (1 - \alpha_j),
\end{equation}

and variances are computed over \(K=5\) neighboring views:

\begin{equation}
    \text{Var}_{\text{views}}(\alpha^i_{\text{eff}, k}) = \frac{1}{K} \sum_{k=1}^K \left( \alpha^i_{\text{eff}, k} - \bar{\alpha}^i_{\text{eff}} \right)^2,
\end{equation}

\begin{equation}
    \text{Var}_{\text{views}}(\mathbf{c}^i(\mathbf{v}_k)) = \frac{1}{K} \sum_{k=1}^K \|\mathbf{c}^i(\mathbf{v}_k) - \bar{\mathbf{c}}^i\|_2^2.
\end{equation}

The physics component \(P^i\) ensures consistency with the underwater model:

\begin{equation}
    P^i = |z^i - \hat{z}^i| + \left| \alpha^i \cdot (1 - e^{-\hat{\beta}^D(\mathbf{x}^i) \cdot z^i}) \right|,
\end{equation}

where \(\hat{z}^i\) is the depth predicted by the Base Rendering Branch for Gaussian \(i\), computed as the distance from the camera to the Gaussian’s center.

PUS is fed into a lightweight MLP \(\phi\) (2-layer, 32 hidden units) to predict pruning probability:

\begin{equation}
    m^i = \sigma(\phi(\text{PUS}^i)),
\end{equation}

where \(\sigma\) is the sigmoid function. Pruning employs Gumbel-Softmax~\cite{jang2016categorical}:

\begin{equation}
    \{\gaussians_{\text{Pruned}}^i\}_{i=1}^{N_{\text{Pruned}}} = \{\gaussians^i \mid m^i < \tau_{\text{adapt}}\},
\end{equation}

with \(\tau_{\text{adapt}}\) as the 95\% of \(m^i\), updated per iteration.

Pruned Gaussians produce \(I_{\text{UR}}^{\text{Enhan.}}\), reducing scattering artifacts compared to \(\hat{I}_{\text{UR}}\). The Physics-Aware Pruning Supervision Loss (PAPSL) is defined in Section~\ref{sec:loss_function}.

\subsection{Learnable Underwater Image Formation Module}

The underwater image formation module simulates light propagation in scattering media, building on Eq.~\eqref{eq.scattering_1}. It utilizes the preliminary Unattenuated Radiance Image (URI) \(\hat{I}_{\text{UR}}\) from the Base Rendering Branch as input to synthesize the underwater image. Integrated with the refinement from the PAUP branch, it replaces intrinsic radiance \(J\) with the enhanced Unattenuated Radiance Image (URI) \(I_{\text{UR}}^{\text{Enhan.}}\), yielding:
\begin{multline}
    I = I_{\text{UR}}^{\text{Enhan.}} \cdot \exp(-\beta^D(\mathbf{v}_D) \cdot z) \\
      + B^{\infty} \cdot \left(1 - \exp(-\beta^B(\mathbf{v}_B) \cdot z)\right),
    \label{eq:scattering_3}
\end{multline}

where unknowns include attenuation coefficients \(\beta^B, \beta^D\), directional dependencies \(\mathbf{v}_B, \mathbf{v}_D\), depth \(z\), and veiling light \(B^{\infty}\). To enable spatial variability while improving efficiency, we adopt a tensor-decomposed voxel grid for regression, guided by PUS from the PAUP branch.

Unlike prior MLP-based approaches~\cite{levy2023seathru, sethuraman2023waternerf}, our method leverages low-rank tensor decomposition inspired by TensoRF~\cite{Chen2022ECCV}, reducing memory and query costs. We adopt:

\begin{itemize}
    \item \textbf{Veiling Light:} \(B^{\infty}\) is a learnable RGB vector \(\hat{B}^{\infty} \in \mathbb{R}^3\).
    \item \textbf{Attenuation Coefficients:} \(\beta^B\) and \(\beta^D\) are regressed using voxel grids \(V^D, V^B \in \mathbb{R}^{G \times G \times G \times 3}\) (\(G=64\)), via vector-matrix (VM) decomposition:
    \begin{equation}
        V^D \approx \sum_{r=1}^R \mathbf{u}_r^D \mathbf{M}_r^D (\mathbf{v}_r^D \circ \mathbf{w}_r^D),
    \end{equation}
    where \(R=16\), \(\mathbf{u}_r^D \in \mathbb{R}^G\), \(\mathbf{M}_r^D \in \mathbb{R}^{G \times G}\), and \(\mathbf{v}_r^D, \mathbf{w}_r^D \in \mathbb{R}^G\). Parameters are queried as \(\hat{\beta}^D(\mathbf{x}) = \text{Query}(V^D, \mathbf{x})\), \(\hat{\beta}^B(\mathbf{x}) = \text{Query}(V^B, \mathbf{x})\) via trilinear interpolation.
    \item \textbf{Depth Estimation:} Depth \(z\) is sourced from Eq.~\eqref{eq:base_depth}.
    \item \textbf{Directional Dependencies:} We assume isotropic media, omitting \(\mathbf{v}_B, \mathbf{v}_D\), to focus on spatial variations.
\end{itemize}

The final rendering equation is:

\begin{multline}
    I_{\text{UW}} = I_{\text{UR}}^{\text{Enhan.}} \cdot \exp(-\hat{\beta}^D(\mathbf{x}) \cdot z) \\
                 + \hat{B}^{\infty} \cdot (1 - \exp(-\hat{\beta}^B(\mathbf{x}) \cdot z)).
    \label{eq:scattering_4}
\end{multline}

\begin{figure*}[h]
\centering
\includegraphics[width=0.9\linewidth]{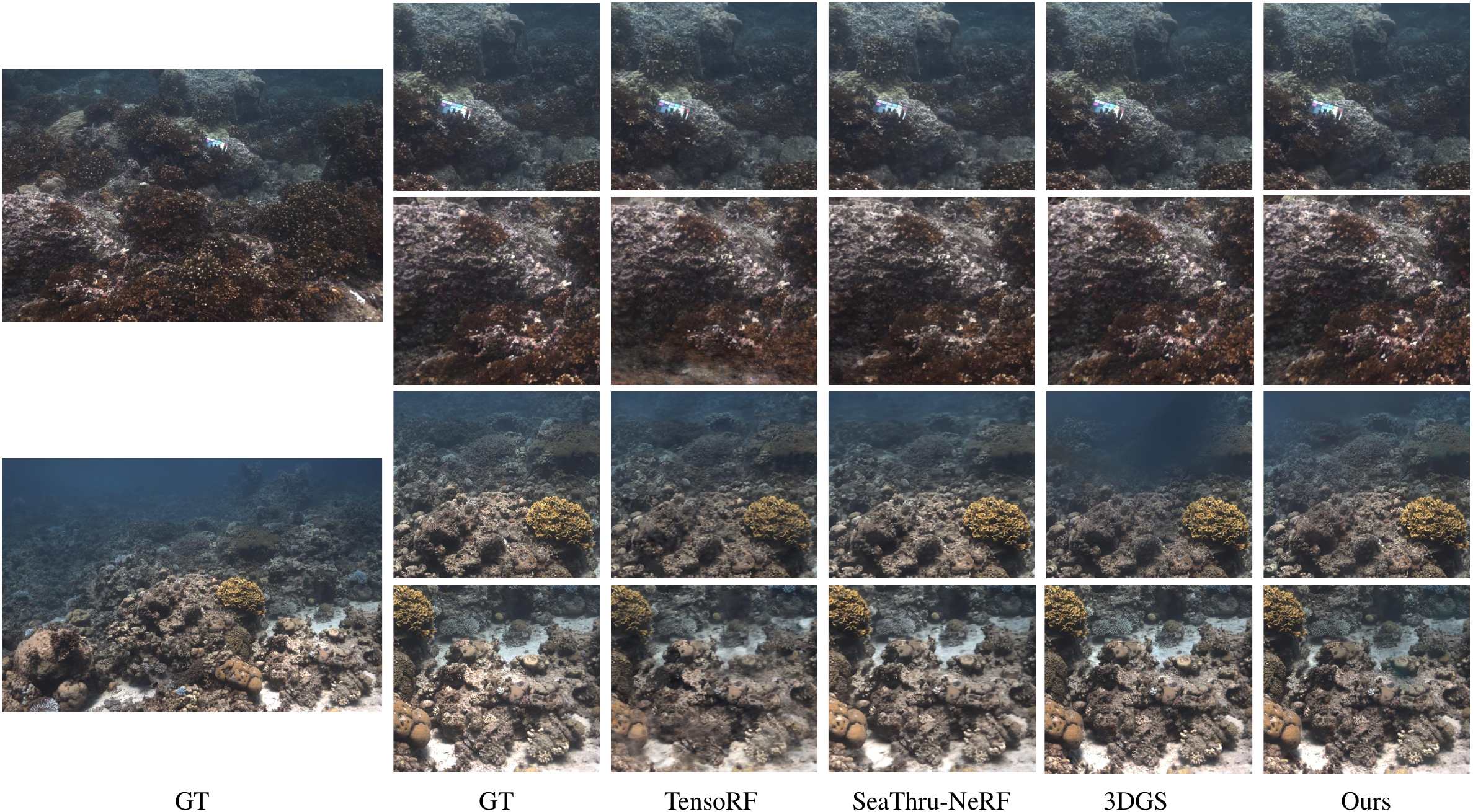}
\caption{Visualization of rendered underwater images.}
\label{fig:unwater_image_demo} 
\end{figure*}

\subsection{Loss Function}
\label{sec:loss_function}

To optimize UW-3DGS, we define a total loss that integrates all components for end-to-end training:
\begin{equation}
    \mathcal{L}_{\text{total}} = \mathcal{L}_{\text{base}} + \lambda_{\text{PAPSL}} \mathcal{L}_{\text{PAPSL}} + \lambda_{\beta} \mathcal{L}_{\beta} + \lambda_{z} \mathcal{L}_{z},
    \label{eq:total_loss}
\end{equation}
where \(\lambda_{\text{PAPSL}} = 0.1\), \(\lambda_{\beta} = 0.05\), and \(\lambda_{z} = 0.05\) balance the contributions of each term. Below, we detail each loss function, its purpose, and its components.

\subsubsection{Image Rendering Loss (\(\mathcal{L}_{\text{IMG}}\))}

The image rendering loss ensures that the rendered underwater image \(I_{\text{UW}}\) matches the ground-truth underwater image \(I_{\text{GT}}\):
\begin{equation}
    \mathcal{L}_{\text{IMG}} = (1 - \lambda) \left\| I_{\text{UW}} - I_{\text{GT}} \right\|_1 + \lambda \mathcal{L}_{\text{D-SSIM}}(I_{\text{UW}}, I_{\text{GT}}),
    \label{eq:img_loss}
\end{equation}
where \(\lambda = 0.2\) balances the L1 loss (pixel-wise intensity difference) and the differentiable SSIM loss (\(\mathcal{L}_{\text{D-SSIM}}\)), which captures structural similarity. The L1 term is computed over all pixels in the image.

\subsubsection{Physics-Aware Pruning Supervision Loss (\(\mathcal{L}_{\text{PAPSL}}\))}

The PAUP branch is optimized with:
\begin{equation}
    \mathcal{L}_{\text{PAPSL}} = \left\| I_{\text{UR}} - I_{\text{UR}}^{\text{Enhan.}} \right\|_1 + \lambda_s \sum_i (1 - m^i) + \lambda_w \|\phi\|_2^2,
    \label{eq:papsl_loss}
\end{equation}
where \(\lambda_s = 0.01\), \(\lambda_w = 0.001\). The first term (L1 over pixels) encourages similarity between unpruned and pruned Unattenuated Radiance Images, reducing scattering artifacts. The second term (sum over Gaussians \(i\)) promotes pruning by penalizing high pruning probabilities \(m^i\). The third term regularizes the MLP \(\phi\) to prevent overfitting.

\subsubsection{Attenuation Regression Loss (\(\mathcal{L}_{\beta}\))}

The attenuation coefficients are regressed with the loss function:
\begin{multline}
    \mathcal{L}_{\beta} = \sum_{\mathbf{x}} \text{PUS}(\mathbf{x}) \cdot \|\hat{\beta}(\mathbf{x}) - \beta_{\text{prior}}\|_2^2 + \\\lambda_r \sum_{r=1}^R \left( \|\mathbf{u}_r^D\|_2^2 + \|\mathbf{v}_r^D\|_2^2 + \|\mathbf{w}_r^D\|_2^2 + \|\mathbf{M}_r^D\|_F^2 \right),
    \label{eq:beta_loss_vm}
\end{multline}

where $\lambda_r = 0.001$, and the first sum is over all voxel positions $\mathbf{x}$. $\beta_{\text{prior}}$ is an empirical mean attenuation coefficient. The PUS term weights the loss to prioritize scattering-dominated regions. A symmetric term applies to $\hat{\beta}^B$. The second term regularizes the \textbf{Vector-Matrix (VM) decomposition components} for the voxel grid $V^D$, penalizing the L2 norm of the vectors ($\mathbf{u}_r^D, \mathbf{v}_r^D, \mathbf{w}_r^D$) and the Frobenius norm of the matrices ($\mathbf{M}_r^D$).

\subsubsection{Depth Refinement Loss (\(\mathcal{L}_{z}\))}

The depth refinement loss is:
\begin{equation}
    \mathcal{L}_{z} = \sum_i (1 - U^i) \cdot |z - \hat{z}^i|,
    \label{eq:z_loss}
\end{equation}
where the sum is over Gaussians \(i\), \(z\) is the rendered depth from Eq.~\eqref{eq:base_depth}, and \(\hat{z}^i\) is the predicted depth for Gaussian \(i\) (distance from the camera to its center). The \((1 - U^i)\) term prioritizes low-uncertainty Gaussians to refine depth estimates.

\begin{figure*}[ht]
\centering

% 子图 (a): 3DGS
\subfigure[3DGS]{
\makecell{
\includegraphics[width=0.18\linewidth, height=0.12\linewidth]{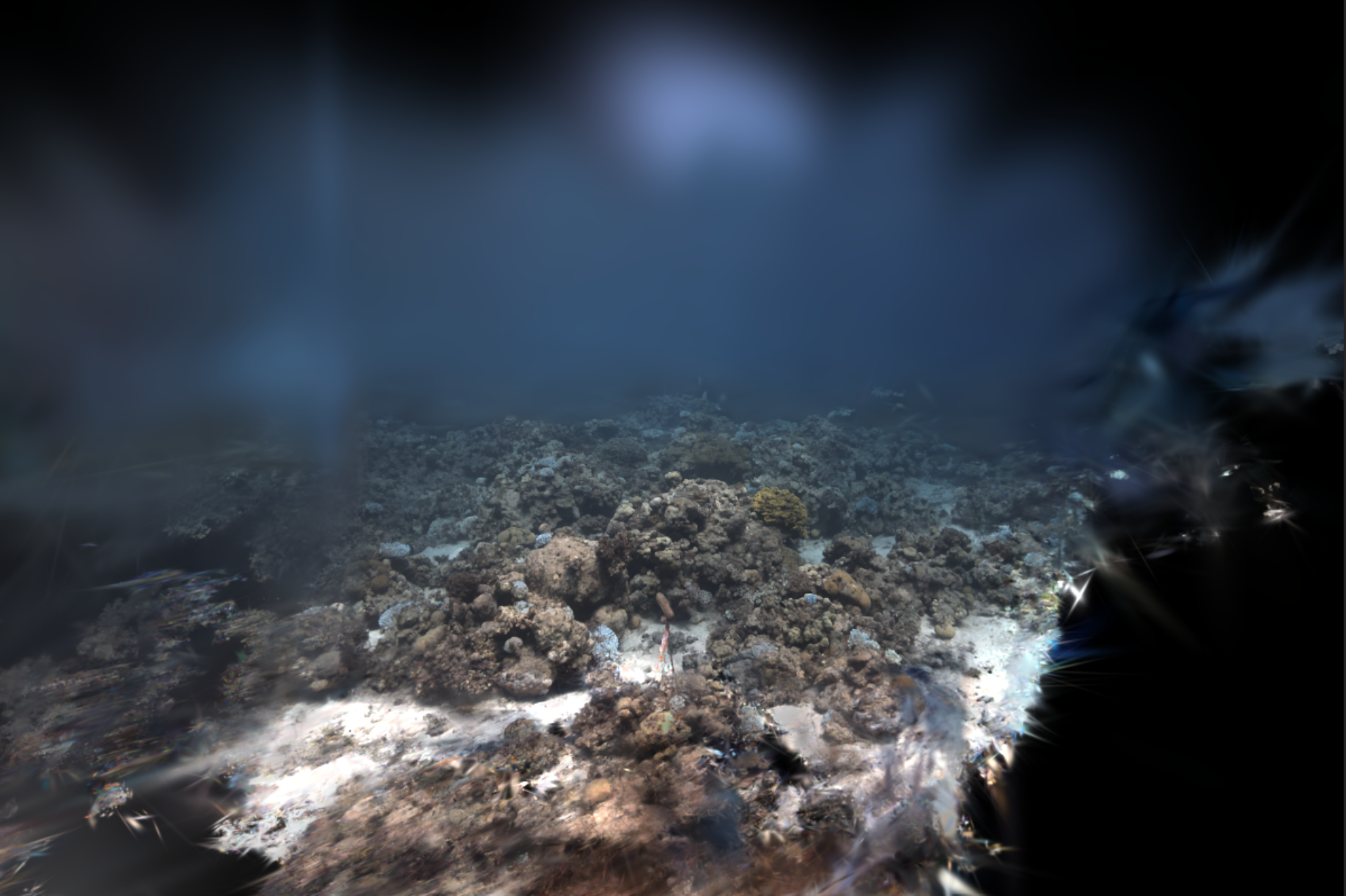}\\
\includegraphics[width=0.18\linewidth, height=0.12\linewidth]{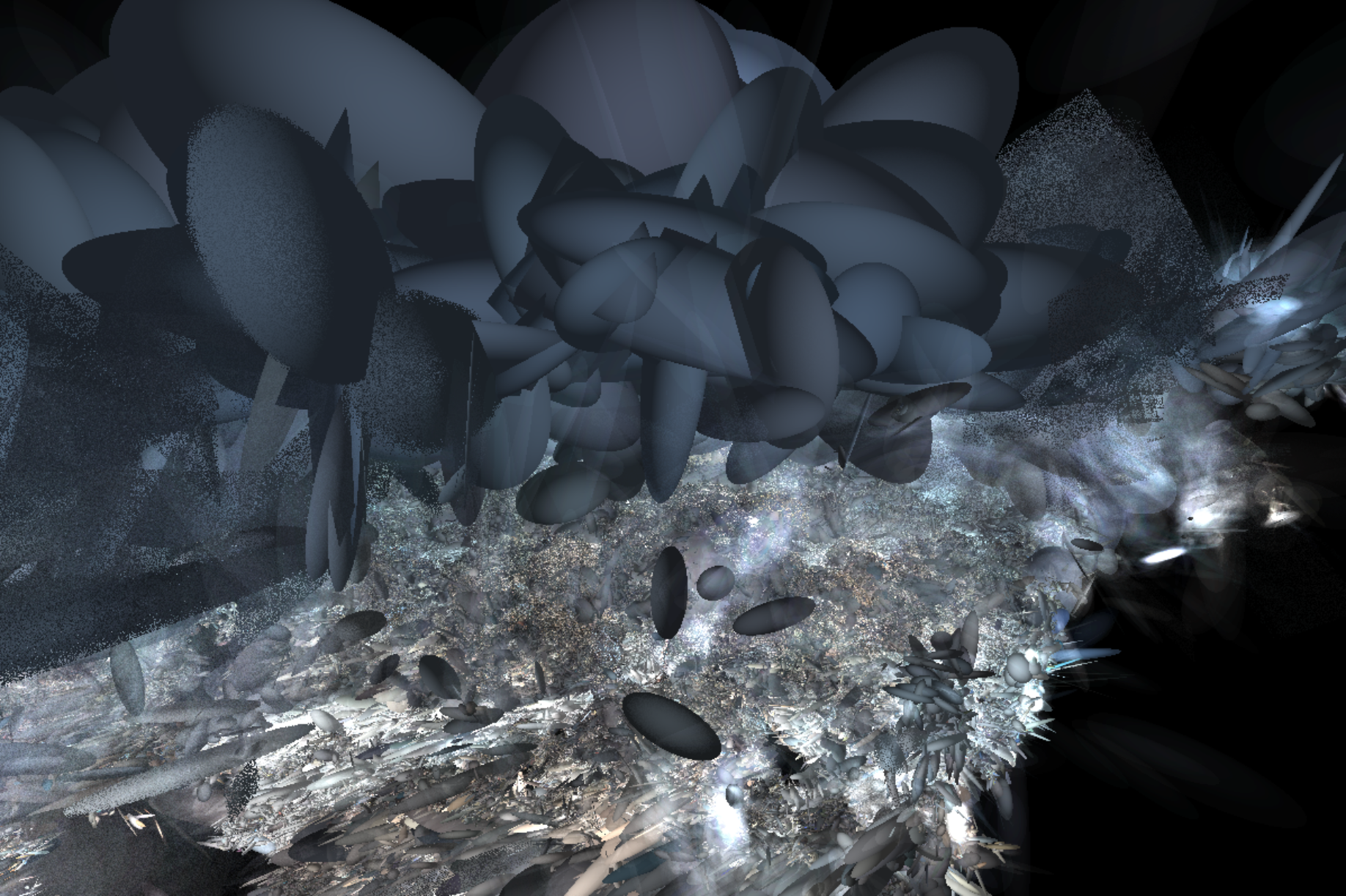}
}
}
\hspace{2em}
\subfigure[Ours - w/ $\mathcal{L}_\text{PAPSL}$]{
\makecell{
% 上两张图
\includegraphics[width=0.18\linewidth, height=0.12\linewidth]{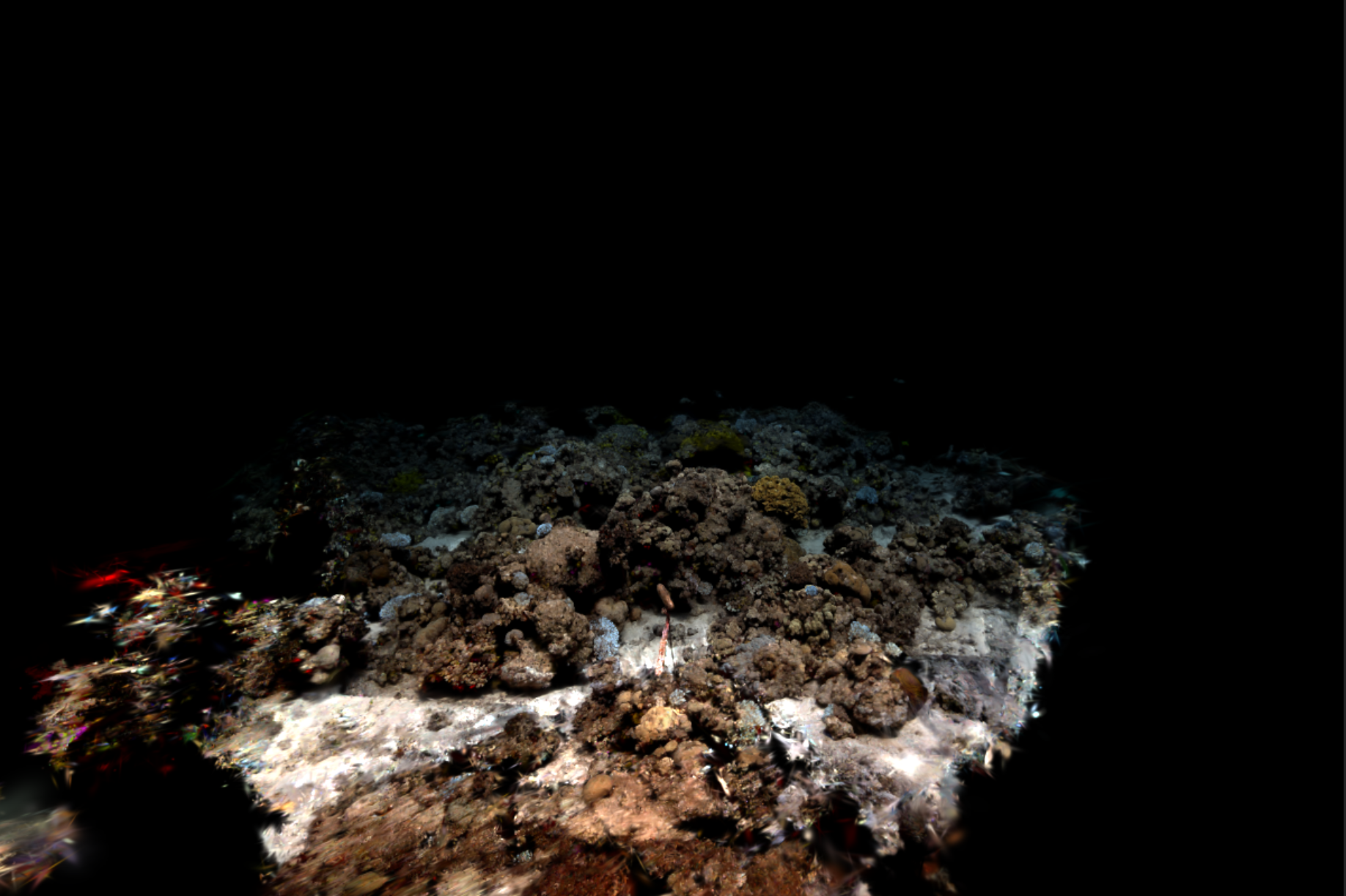}
\includegraphics[width=0.18\linewidth, height=0.12\linewidth]{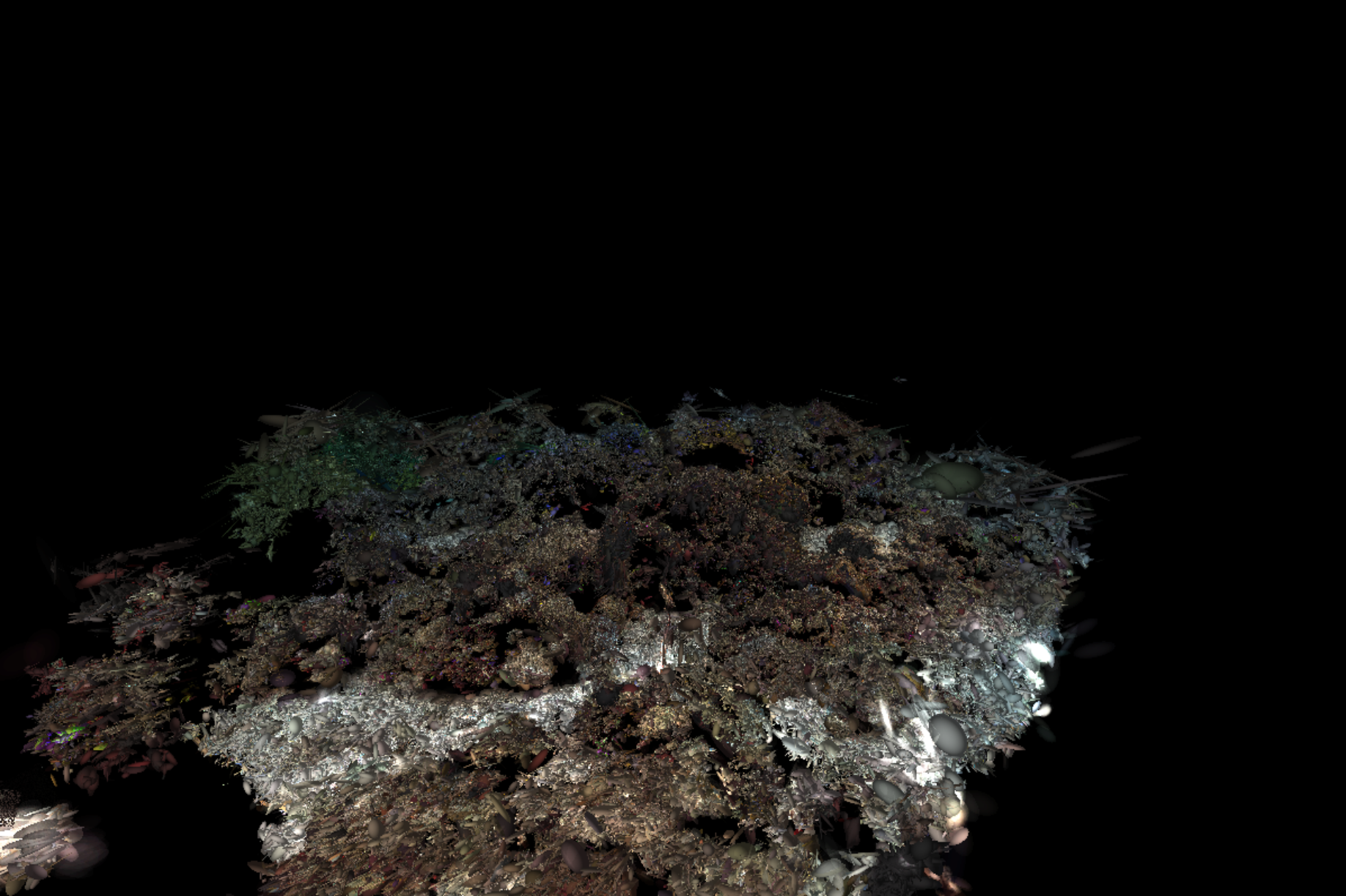}
\includegraphics[width=0.18\linewidth, height=0.12\linewidth]{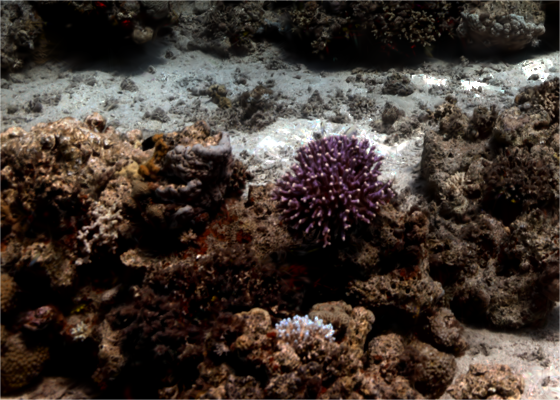}
\includegraphics[width=0.18\linewidth, height=0.12\linewidth]{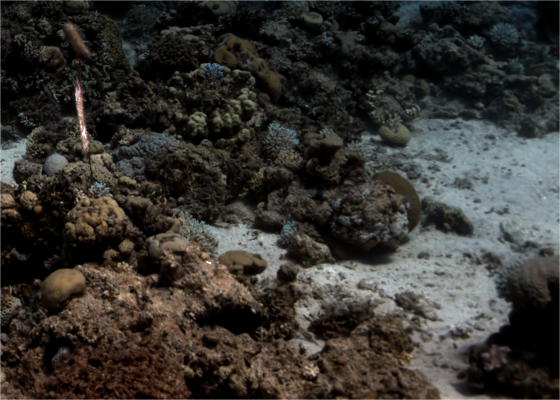}\\
\includegraphics[width=0.18\linewidth, height=0.12\linewidth]{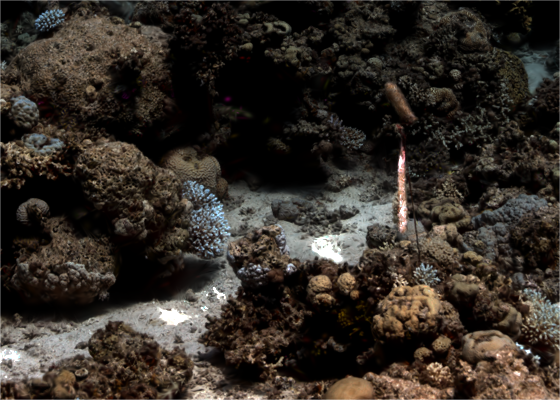}
\includegraphics[width=0.18\linewidth, height=0.12\linewidth]{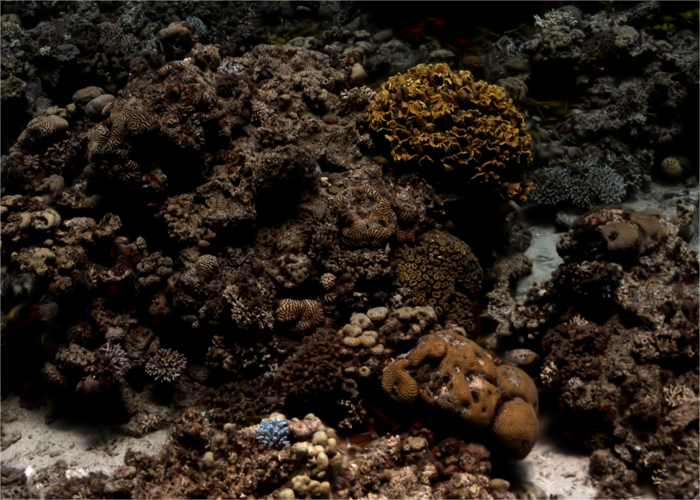}
\includegraphics[width=0.18\linewidth, height=0.12\linewidth]{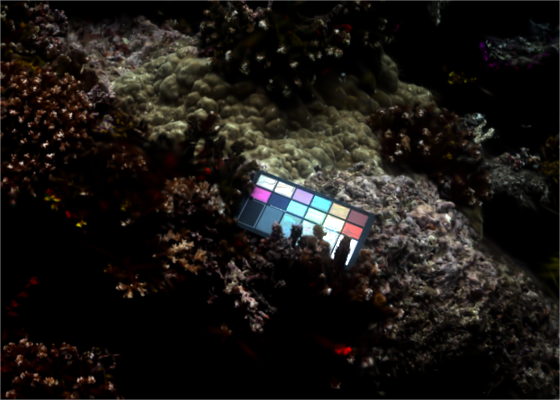}
\includegraphics[width=0.18\linewidth, height=0.12\linewidth]{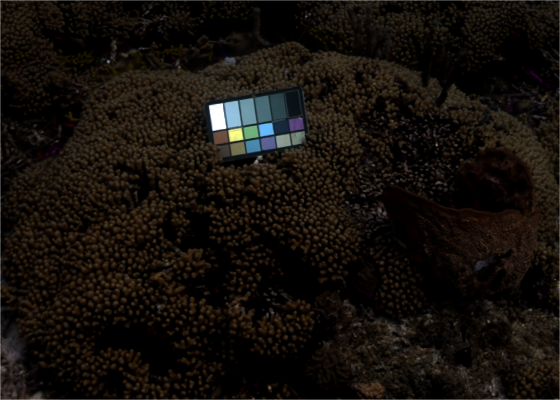}
}
}

\caption{Unattenuated Radiance Images (URI) of seabed, directly rendered from 3D Gaussians trained with the Physics-Aware Pruning Supervision Loss (PAPSL), demonstrating effective suppression of floating artifacts.}
\label{fig:combined_underwater_visualization}
\end{figure*}

\section{Implementation Details}
\label{sec:implementation}

UW-3DGS is implemented in PyTorch with CUDA, based on 3D Gaussian Splatting (3DGS)~\cite{kerbl3Dgaussians}, and trained on a single NVIDIA Tesla V100 GPU (32\,GB). Training runs for 40,000 iterations using the Adam optimizer~\cite{kingma2014adam} with a batch size of one image. Key configurations include:

\begin{itemize}
    \item {Hyperparameters}: Learning rates are 0.05 (opacity), 0.005 (scaling), 0.001 (rotation), and 0.001 (\(\hat{\beta}^B\), \(\hat{\beta}^D\), \(\hat{B}^{\infty}\)). Loss weights: \(\lambda_{\text{PAPSL}} = 0.1\), \(\lambda_{\beta} = \lambda_{z} = 0.05\), \(\lambda = 0.2\) (\(\mathcal{L}_{\text{IMG}}\)), \(\lambda_{\beta} = 0.01\) (\(\mathcal{L}_{\text{base}}\)), \(\lambda_s = 0.01\), \(\lambda_w = \lambda_r = 0.001\). Voxel grid resolution is \(G=64\), rank \(R=16\). PAUP parameters: \(w_u = w_p = 0.5\), \(w_\alpha = 0.4\), \(w_c = 0.6\), \(K=5\). MLP \(\phi\): 2 layers, 32 units. 
    \item {Training Setup}: Gaussian densification starts at iteration 500 (rate 0.01), with opacity resets every 3,000 iterations. The PAUP branch activates at iteration 500, with pruning threshold \(\tau_{\text{adapt}}\) as the median of pruning probabilities \(m^i\), updated per iteration. Spherical harmonics are truncated at order three.
    \item {Preprocessing}: SeaThru-NeRF images~\cite{levy2023seathru} are white-balanced; UWBundle images~\cite{Skinner:2016ab} use raw data. Initial Gaussians come from COLMAP~\cite{schoenberger2016sfm}. Images are resized to 1024$\times$1024. \(\beta_{\text{prior}}\) is set to \([0.1, 0.15, 0.2]\) (RGB) from dataset statistics.
\end{itemize}

\section{Experiments}
This section introduces the experiment settings and results.
 All experimental results are obtained through our rerunning.

\subsection{Datasets}
\label{sec:datasets}
We evaluate UW-3DGS on \textbf{UWBundle}~\cite{Skinner:2016ab} and \textbf{SeaThru-NeRF}~\cite{levy2023seathru} datasets, covering synthetic and real-world underwater scenarios. UWBundle has 36 synthetic images of a submerged rock platform, captured in a lab with a lawnmower trajectory. SeaThru-NeRF includes 58 white-balanced images from the Pacific (Panama), Red Sea (Israel), and Caribbean (Curaçao), with challenges like variable water properties. Official training/testing splits from~\cite{mildenhall2019llff, levy2023seathru} ensure fair comparisons.

\subsection{Evaluation Rubrics}
\label{sec:eva_rubics}
We assess UW-3DGS on three aspects: rendering quality of underwater (UWI) and no-water (URI) images, training efficiency, and 3D reconstruction quality without water. Rendering quality is measured against ground-truth images using PSNR ($\uparrow$), SSIM ($\uparrow$), and LPIPS ($\downarrow$)~\cite{zhang2018unreasonable}.

\subsection{Competing Methods}
\label{sec:competing_methods}

We compare UW-3DGS with key methods for underwater 3D reconstruction, focusing on rendering quality and efficiency. These methods include
3DGS \cite{kerbl3Dgaussians}, TensoRF \cite{Chen2022ECCV}, SeaThru-NeRF \cite{levy2023seathru}, WaterSplatting\cite{li2024watersplatting}, SeaSplat \cite{yang2024seasplat}.

\subsection{Underwater Image Rendering Quality Comparison}

Table~\ref{tab:quantative_virtual_views} presents the quantitative evaluation of novel view synthesis quality for underwater images (UWIs) on the SeaThru-NeRF dataset. Among all variants, our method performs favorably on UWI rendering tasks, achieving the best SSIM among compared methods. These results confirm that modeling underwater light attenuation and scattering significantly improves photorealistic rendering under aquatic conditions.

\begin{table}[h]
\centering
\caption{Quantitative comparisons of underwater image rendering quality averaged on the SeaThru-NeRF dataset.}
\small
\begin{tabular}{l c c c}
\toprule
\multirow{2}{*}{Method} & \multicolumn{3}{c}{Metric} \\
\cmidrule(lr){2-4}
& PSNR$\uparrow$ & SSIM$\uparrow$ & LPIPS$\downarrow$ \\
\midrule
3DGS & 26.113 & 0.861 & 0.216 \\
TensoRF & 24.307 & 0.787 & 0.285 \\
SeaThru-NeRF & 25.768 & 0.806 & - \\
WaterSplatting & \textbf{29.687} & {0.830} & \textbf{0.120} \\
SeaSplat  & 27.108 & 0.835 & 0.183 \\
Ours & 27.604 & \textbf{0.868} & 0.134 \\
\bottomrule
\end{tabular}
\label{tab:quantative_virtual_views}
\end{table}

\begin{table*}[h]
    \centering
    \caption{Ablation study results on SeaThru-NeRF dataset (averaged across scenes).}
    \small
    \begin{tabular}{l c c c c c}
    \toprule
    Variant & PSNR$\uparrow$ & SSIM$\uparrow$ & LPIPS$\downarrow$ & Train Time (min) & Big Float Gauss Ratio (\%) $\downarrow$\\
    \midrule
    w/o PAUP & 25.374 & 0.837 & 0.266 & 42 & 8.2 \\
    w/o Underwater Module & 24.912 & 0.812 & 0.298 & 28 & 6.5 \\
    w/o \(\mathcal{L}_{\beta}\) & 25.754 & 0.851 & 0.247 & 46 & 5.8 \\
    w/o \(\mathcal{L}_{z}\) & 26.028 & 0.859 & 0.266 & 47 & 7.4 \\
    Full UW-3DGS & \textbf{27.604} & \textbf{0.868} & \textbf{0.134} & 48 & \textbf{1.3} \\
    \bottomrule
    \end{tabular}
    \label{tab:ablation_results}
\end{table*}

\subsection{Unattenuated Radiance Image Rendering Quality Comparisons}
The visualization results in Figure~\ref{fig:combined_underwater_visualization} demonstrate the effectiveness of our method trained with the Physics-Aware Pruning Supervision Loss (PAPSL), producing clear Unattenuated Radiance Images (URI) of the seabed by directly rendering 3D Gaussians. Compared to standard 3DGS, which exhibits noisy floating Gaussians and blurred topography, UW-3DGS yields sharper geometric details, such as well-defined coral formations and marine flora, with minimal artifacts, highlighting PAPSL's role in suppressing scattering-induced noise.
Furthermore, Figure~\ref{fig:combined_underwater_visualization} shows URI comparisons between UW-3DGS and SeaThru-NeRF on the SeaThru-NeRF dataset. Our method reconstructs underwater scenes with superior clarity, preserving intricate seabed contours and reducing volumetric haze, leading to more accurate and visually coherent results. This underscores UW-3DGS's advantage in disentangling scattering effects, resulting in better overall underwater reconstruction quality, including enhanced depth accuracy and artifact-free geometry, essential for applications like marine exploration.

To validate the contributions of UW-3DGS's key components—Base Rendering Branch, Physics-Aware Uncertainty Pruning (PAUP) Branch, and Learnable Underwater Image Formation Module—we conduct ablation studies on the SeaThru-NeRF dataset. We evaluate variants by removing or modifying components and assess impacts on rendering quality (PSNR, SSIM, LPIPS), training time, and geometric fidelity (measured by floating Gaussian ratio, i.e., percentage of pruned Gaussians). Details will be discussed in the subsequent section.

\subsection{Ablation Study}
\label{sec:abla}

We test the following variants:
\begin{itemize}
    \item \textbf{w/o PAUP}: Disables the PAUP Branch and \(\mathcal{L}_{\text{PAPSL}}\), relying only on base rendering and underwater module.
    
    \item\textbf{w/o Underwater Module}: Removes the learnable underwater model (Eq.~\eqref{eq:scattering_4}) and associated losses (\(\mathcal{L}_{\beta}\), \(\mathcal{L}_{z}\)), using standard 3DGS rendering.
    \item\textbf{w/o \(\mathcal{L}_{\beta}\)}: Omits attenuation regression loss, using fixed \(\beta_{\text{prior}}\) instead of learned coefficients.
    \item \textbf{w/o \(\mathcal{L}_{z}\)}: Disables depth refinement loss, using unweighted depth computation.
    \item \textbf{Full UW-3DGS}: Complete method with all components.
\end{itemize}

Results are in Table~\ref{tab:ablation_results}, the full UW-3DGS model achieves the highest rendering quality (PSNR: 27.604, SSIM: 0.868, LPIPS: 0.134) and the lowest big floating Gaussian ratio (1.3\%), demonstrating superior geometric fidelity and artifact reduction. Ablations reveal that removing PAUP markedly increases the ratio to 8.2\% and degrades PSNR by over 2 dB, underscoring its role in pruning noisy Gaussians. Similarly, omitting the underwater module or specific losses elevates artifacts and lowers performance, confirming the synergistic necessity of all components for balanced efficiency and fidelity.

\section{Conclusions}
We propose UW-3DGS, an efficient framework for underwater 3D scene reconstruction that integrates a physically grounded image formation model into the 3D Gaussian Splatting pipeline. This enables simultaneous geometry recovery and color restoration, producing high-fidelity renderings of both underwater and media-free appearances.
To address scattering artifacts, we introduce a Physics-Aware Uncertainty Pruning Branch, which refines noisy Gaussians and yields clean, physically consistent reconstructions. UW-3DGS excels in media-free rendering by generating clear radiance images and depth maps, preserving fine details such as coral textures and seabed structures—critical for marine ecology and robotic perception.
Experiments on the SeaThru-NeRF dataset demonstrate superior rendering quality and geometric accuracy. UW-3DGS offers a promising solution for underwater exploration, marine robotics, and environmental monitoring.

\section{Limitations}
(1) The fixed voxel grid resolution with tensor decomposition may inadequately capture fine spatial variations in large-scale scenes, especially in environments with significant depth changes. (2) The PAUP branch's uncertainty computation relies on variance over neighboring views, potentially reducing robustness in sparse viewpoint scenarios common in underwater data collection.

\bibliography{aaai}

\end{document}